\documentclass[runningheads]{llncs}
\usepackage[T1]{fontenc}
\usepackage{graphicx}
\usepackage{booktabs}
\usepackage[misc]{ifsym}
\usepackage{tabularx,array}
\newcolumntype{L}{>{\raggedright\arraybackslash}X}
\newcolumntype{C}{>{\centering\arraybackslash}X}

\newcommand{\EpsTot}{\varepsilon_{\text{total}}}
\newcommand{\RDP}{\mathrm{RDP}}

\usepackage{float}
\usepackage{mwe}
\usepackage{amsmath,amssymb}
\usepackage{algorithm}
\usepackage{algpseudocode}
\usepackage{tikz}
\usepackage{hyperref} 

\begin{document}

\title{Adaptive Token-Weighted Differential Privacy for LLMs: Not All Tokens Require Equal Protection}
\titlerunning{Adaptive Token-Weighted Differential Privacy for LLMs} 

\author{Manjiang Yu\inst{1} \and Priyanka Singh\inst{1} \and Xue Li\inst{1} \and Yang Cao\inst{2}}
\authorrunning{Yu, Singh, Li \& Cao}    

\institute{The University of Queensland\\
\email{\{manjiang.yu, priyanka.singh, xueli\}@uq.edu.au}
\and
Institute of Science Tokyo\\
\email{cao@c.titech.ac.jp}
}

\maketitle

\begin{abstract}
Large language models (LLMs) frequently memorize sensitive or personal information, raising significant privacy concerns. Existing variants of differential privacy stochastic gradient descent (DP-SGD) inject uniform noise into every gradient step, significantly extending training time and reducing model accuracy. We propose that concentrating noise primarily on gradients associated with sensitive tokens can substantially decrease DP training time, strengthen the protection of sensitive information, and simultaneously preserve the model’s performance on non-sensitive data. We operationalize this insight through \textbf{Adaptive Token-Weighted Differential Privacy (ATDP)}, a modification of vanilla DP-SGD that adaptively assigns different gradient weights to sensitive and non-sensitive tokens. By employing a larger noise scale at the early stage of training, ATDP rapidly disrupts memorization of sensitive content. As a result, ATDP only requires a few additional epochs of lightweight post-processing following standard fine-tuning, injecting targeted noise primarily on parameters corresponding to sensitive tokens, thus minimally affecting the model’s general capabilities. ATDP can be seamlessly integrated into any existing DP-based fine-tuning pipeline or directly applied to non-private models as a fast privacy-enhancing measure. Additionally, combined with an initial redacted fine-tuning phase, ATDP forms a streamlined DP pipeline that achieves comparable canary protection to state-of-the-art DP-SGD methods, significantly reduces the computational overhead of DP fine-tuning, shortening training time by approximately 90\% , while achieving comparable or superior privacy protection and minimal accuracy degradation.

\end{abstract}
\begin{tikzpicture}[remember picture,overlay]
  \node[anchor=south west, xshift=1cm, yshift=1cm] 
      at (current page.south west) 
      {\footnotesize Our code and data are available at \url{https://github.com/1784035059/ATDP}};
\end{tikzpicture}
\section{Introduction}
Large language models (LLMs) play a critical role in natural language processing and have found widespread applications. Unfortunately, \cite{carlini2022quantifying}while LLMs process enormous amounts of text data in real-world scenarios, they also inevitably memorize certain portions of their training data, which may include private or sensitive information about individual users. More concerning is that, due to this memory characteristic, LLMs may inadvertently leak sensitive information (such as names, ID numbers, passwords, or medical data) when generating text. This poses significant ethical and legal risks, making research on privacy protection for language models particularly important. Differential Privacy (DP)\cite{dwork2008differential} is one of the most theoretically well-founded methods for privacy protection. By limiting how training data influence model parameters, it ensures that adversaries find it difficult to reconstruct the original data. However, applying traditional DP-based fine-tuning (such as DPSGD\cite{abadi2016deep}) to large models often leads to significant performance degradation. Previous work has attempted to address this issue by applying DP in various ways, but these approaches typically apply noise uniformly to all samples or all tokens, without differentiation. Since sensitive information in language data is often sparse, adding high-intensity noise to every token results in unnecessary performance loss and training instability. Even attempts that selectively handle different tokens have not yet achieved the fine-grained single-token-level control that is needed.

A key fact is that the proportion of sensitive tokens may vary across different datasets, but it almost always constitutes only a small fraction of the data. Meanwhile, the vast majority of ordinary tokens do not contain private information; allowing the model to learn these normally helps maintain strong language modeling performance. Therefore, if we can differentiate between sensitive and non-sensitive tokens when applying noise, there is an opportunity to minimize the impact on the model’s core capabilities while still effectively protecting sensitive information. However, in practice, it remains challenging to determine which tokens are sensitive and how to adaptively adjust the noise intensity during training.

LLMs, popularized by Vaswani et al.~\cite{vaswani2017attention}, have demonstrated impressive success in the field of natural language processing. Trained on vast quantities of public text data, these models excel at capturing general linguistic structures. In practice, a common approach is to fine-tune these pre-trained LLMs on specific downstream tasks, this method remains effective even when privacy constraints are present during training. Previously, Shi et al.\cite{shi2022just}proposed a two-stage fine-tuning approach that edits out sensitive tokens from the dataset and then fine-tunes the model twice—once with the edited data and once with the original data. This method showed strong results in both protecting private information and maintaining model accuracy; however, their second stage injects identical noise into each gradient step, significantly prolonging training time.
We propose \emph{Adaptive Token-Weighted Differential Privacy (ATDP)}, which concentrates the differential privacy noise primarily on gradients corresponding to sensitive tokens. By adaptively weighting token gradients, ATDP significantly reduces DP training time, offers stronger protection of sensitive information, and preserves most of the model's original performance on non-sensitive data. As a result, ATDP efficiently suppresses memorization of sensitive content while minimally affecting the overall model quality.

ATDP can be seamlessly integrated as a final step into any DP-SGD fine-tuning process, injecting targeted noise to significantly reduce canary exposure without materially enlarging the composed \((\varepsilon,\delta)\) at our reporting precision. We report the composed privacy loss over all stages and keep fixed \(\varepsilon\) across ablations. It can also be directly applied to a non-private model, effectively suppressing the memorization of sensitive tokens. Crucially, when combined with a single redacted fine-tuning stage, ATDP provides a streamlined DP pipeline that matches or surpasses the privacy protection of a strong DP-SGD baseline, while dramatically reducing training time by approximately 90\% and incurring only minimal loss in downstream accuracy. This substantial improvement in computational efficiency and the preservation of model performance represent the key contributions of this work.

\section{Related Work}

LLMs tend to memorize portions of their training data~\cite{carlini2019secret}. For instance, Carlini et al.~\cite{carlini2021extracting,nasr2023scalable} managed to extract gigabytes of training data including sensitive information, from various models: open-source ones like Pythia~\cite{biderman2023pythia} and GPT-Neo~\cite{black2022gpt}, semi-open models such as LLaMA~\cite{touvron2023llama} and Falcon, and even closed-source services like ChatGPT.
DP~\cite{dwork2008differential,dwork2014algorithmic} is widely used to protect personal information in data. Abadi et al.~\cite{abadi2016deep} introduced the DP-SGD algorithm for training deep learning models. Although DP-SGD lowers the risk of leaking information from training data, adding random noise to gradients often hurts model performance~\cite{li2021large} and can even prevent large models from converging.

To address this issue, Li et al.~\cite{li2021large} explored large pre-trained language models, experimented with non-standard hyperparameters better suited for DP optimization, and adopted fine-tuning objectives aligned with the original pre-training scheme. Many studies have also explored adding different noise levels in differential privacy fine-tuning according to the varying sensitivity of the data. Shi et al.~\cite{shi2021selective} introduce Just Fine-tune Twice (JFT), a two-step approach to achieve Selective Differential Privacy (SDP) for LLMs. First, they fine-tune on redacted data to capture in-domain patterns, then privately fine-tune again on the original data. Wu et al.~\cite{wu2022adaptive} propose Adaptive Differential Privacy (ADP) for language modeling, which automatically gauges each token’s privacy probability via perplexity – assuming rare text is more likely private. The method then adjusts the DP noise for batches with higher privacy scores, thus preserving utility while still limiting leakage risk. Wang et al.~\cite{wang2024selective} introduce a new privacy concept called Selective Sequence Local Differential Privacy (S-SeqLDP), which injects noise only into the sensitive segments of embeddings. Their system, SLDP-FT, uses a “privacy forward weight” to determine the noise magnitude for each token, providing targeted protection while maintaining a theoretically proven level of privacy. There are also many other works~\cite{tramer2020differentially,papernot2016semi,ghazi2021deep,yu2021differentially} employ public data for private learning, but unlike these approaches, our method is the first to introduce dynamic token-level loss weighting in differential privacy, enabling the model to treat data of varying sensitivity differently.

\section{Preliminary}
\subsection{Differential Privacy}
Differential Privacy (DP) is a formal framework that protects individual information in a dataset.  
Intuitively, an algorithm is said to be $(\epsilon,\delta)$-DP if observing its output does not allow an adversary to determine the presence or absence of any single record by more than a factor of $e^\epsilon$. Formally, for any pair of datasets $D_1$ and $D_2$ that differ in exactly one record, and for any set $S$ in the algorithm's output space, a randomized algorithm $A$ satisfies $(\epsilon,\delta)$-DP if:
\[
  \Pr\bigl(A(D_1) \in S\bigr)
  \;\le\; 
  e^\epsilon \,\Pr\bigl(A(D_2) \in S\bigr) + \delta.
\]
where smaller values of $\epsilon$ imply stronger privacy protection.

\subsection{DP-SGD}

Differentially Private Stochastic Gradient Descent (DP-SGD)\cite{abadi2016deep} is a widely used method to train machine learning models with rigorous privacy guarantees. In essence, it introduces two main modifications to standard SGD: gradient clipping and noise addition. Specifically, given a dataset $D$ and model parameters $\theta$, at each iteration, a minibatch $B$ of size $m$ is sampled, and the per-example gradients $\{g_i\}$ for each $x_i \in B$ are computed. Each gradient $g_i$ is then clipped so that its $\ell_2$ norm does not exceed a predefined bound $C$, ensuring that no single record can disproportionately affect the parameter updates. The clipped gradients $\{\tilde{g}_i\}$ are averaged to form $\bar{g}$, and isotropic Gaussian noise $\mathcal{N}(0, \sigma^2 C^2 I)$ is added to achieve differential privacy under $(\epsilon, \delta)$ constraints. Finally, the model parameters are updated via $\theta \leftarrow \theta - \eta \bar{g}$, where $\eta$ is the learning rate. By carefully selecting the noise scale $\sigma$, clipping bound $C$, and other hyperparameters, DP-SGD can limit the overall privacy loss across multiple training iterations. This privacy loss is commonly tracked using the moments accountant or other composition theorems. While DP-SGD helps prevent unintended memorization of sensitive data, it can also degrade model performance if the noise level is too high or if hyperparameters are poorly tuned. Nonetheless, for a broad range of tasks and model sizes, DP-SGD provides a practical balance between data protection and predictive accuracy.

\subsection{Adaptive Token-Weighted Differential Privacy (ATDP)}
 
In this work, we introduce \emph{Adaptive Token-Weighted Differential Privacy (ATDP)}, which still relies on the core principles of DP-SGD (clip-and-noise), but selectively adjusts how much each token is affected by noise. 
it keeps the clip-and-noise core of vanilla DP-SGD but scales each token-level gradient by a fixed weight that is chosen once before training.  
A \emph{secret detector} first marks tokens that may reveal private information.  
During the short ATDP run, gradients for these marked tokens keep their full weight, whereas gradients for all other tokens are multiplied by a near-zero factor.  
A relatively large noise scale is applied from the very first update, so the added noise mainly corrupts the parameters that store sensitive content while leaving the rest of the model almost untouched.  
This selective corruption quickly weakens memorization of private strings yet preserves the overall language modeling ability at the same privacy budget.  
The next section details the algorithm and its formal analysis.

\section{Methodology}

In order to selectively protect sensitive information without overly penalizing general language modeling performance, in this section, we propose \emph{Adaptive Token-Weighted Differential Privacy (ATDP)}. Unlike standard DP-SGD, which applies the same clipping and noise injection to \emph{all} tokens, ATDP focuses noise addition primarily on sensitive tokens. Concretely, before training, we employ a \emph{secret detector} to identify and edit sensitive tokens in the dataset. 

\subsection{Secret Detector}\label{sec:secret_detector}

Typically, \textit{personally identifiable information (PII)} refers to details such as an individual’s full name or Social Security number. However, in the context of NLP, a large portion of private information is \textit{contextual}\cite{brown2022does}. For example, consider the sentence:
\textit{"What approach do you plan to take regarding your upcoming surgery?"}
Although no explicit PII appears, the overall meaning suggests highly sensitive information.

To detect and label such privacy-sensitive content, we adopt the \textit{secret detector} approach from Shi \cite{shi2022just}, which uses a dependency parser and POS tagger in \textsc{spaCy} \cite{honnibal2020spacy} to identify privacy-related tokens in the training data. We use off-the-shelf \textbf{POS} (part-of-speech) tags to mark token categories (noun, verb,
pronoun, etc.) and a \textbf{dependency parser} to obtain grammatical roles (subject, object,
modifier). These signals define our four tiers: low/high-entity rely on pattern matching
and NER; low/high-contextual additionally include roles (and, for the highest tier, predicates).
Based on these, we apply different levels of privacy protection. An illustrative example of how these secret detectors operate and flag potential sensitive content is given below.

\noindent We adopt a four-tier approach that progressively expands from \textit{entity-based} to \textit{contextual} detection:

\begin{enumerate} \item \textbf{Low Entity} focuses on four PII-related categories—\textit{person}, \textit{organization}, \textit{date}, and \textit{location}—identified via spaCy’s NER\cite{honnibal2020spacy}. \item \textbf{High Entity} extends coverage to all 18 entity types recognized by spaCy\cite{honnibal2020spacy}. \item \textbf{Low Contextual} adds proper nouns, pronouns, and sentence subjects/objects on top of the 18 entities. \item \textbf{High Contextual} further incorporates verbs into the low-contextual scope, yielding the most comprehensive coverage. \end{enumerate}

\noindent As the privacy level increases, the set of tokens that are redacted also becomes larger.

\subsection{Manual \& LLM Validation}

Although the rule-based secret detector provides the first pass of privacy tagging, we add two lightweight verification steps to avoid over-relying on its accuracy.

\paragraph{Human audit.}
For each training epoch we draw a stratified $1\%$ sample of sentences—half from those tagged as sensitive, half from those left untagged.  
Annotators review these sentences and correct false positives or false negatives.  
The corrections update a small whitelist/blacklist that is re-applied before the next epoch, so systematic errors do not accumulate. The procedure requires no relabeling of the full corpus, no extra training runs,
and no task‑specific annotations; reviewers only tick a short checklist and
flag edge cases. We also mask potentially identifying spans before review.

\paragraph{LLM cross-check.}
Each sampled sentence is also sent to a state-of-the-art large language model through the OpenAI API (GPT-4o).  
The model flags tokens it judges privacy-relevant; any disagreement with the rule-based tags is either escalated for manual inspection or, by default, the entire sentence is kept in the sensitive set.  
Because only a small sample is processed, this extra pass adds minimal runtime while catching edge cases that the rule engine misses.

While we currently incorporate a limited manual review step to improve detection accuracy during research, a fully automated detection mechanism would be preferable for real-world deployment.

\begin{table}[h]
    \centering
    \footnotesize
    \renewcommand{\arraystretch}{1.3}
    \begin{tabularx}{\columnwidth}{l|X|X}
        \toprule
        \textbf{Secret Detector} 
        & \textbf{Have you finalized the settlement for the Johnson Estate in Berlin?} 
        & \textbf{Is it true that Emma had a medical procedure at St.\ Mary's Hospital?} \\
        \midrule
        Low entity 
        & Have you finalized the settlement for the Johnson Estate in Berlin? 
        & Is it true that Emma had a medical procedure at St.\ Mary's Hospital? \\
        \midrule
        High entity 
        & Have you finalized the settlement for the \texttt{<ORG>} in \texttt{<LOC>}? 
        & Is it true that \texttt{<PERSON>} had a medical procedure at \texttt{<ORG>}? \\
        \midrule
        Low contextual 
        & \texttt{<PRON>} finalized the \texttt{<OBJ>} for the \texttt{<ORG>} in \texttt{<LOC>}? 
        & Is it true that \texttt{<PERSON>} had a \texttt{<OBJ>} at \texttt{<ORG>}? \\
        \midrule
        High contextual 
        & \texttt{<PRON>} \texttt{<VERB>} the \texttt{<OBJ>} for the \texttt{<ORG>} in \texttt{<LOC>}? 
        & Is it true that \texttt{<PERSON>} \texttt{<VERB>} a \texttt{<OBJ>} at \texttt{<ORG>}? \\
        \bottomrule
    \end{tabularx}
    \caption{Outputs of different detector tiers on two sentences that contain named entities. 
    Low-entity masks only canonical PII (none present here), so texts remain unchanged; 
    High-entity additionally masks PERSON/ORG/LOC; 
    Low-/High-contextual further abstract grammatical roles and, for the latter, verbs.}
    \label{tab:secret_detection}
\end{table}

\subsection{Fine-tuning with Adaptive Token-Weighted Differential Privacy}\label{sec:w_non}
Having identified which tokens are sensitive via our secret detector, we now describe the
core of ATDP. We \emph{reweight token losses within each record} before the usual
\emph{per-record clipping + Gaussian noise} step of DP-SGD. Sensitive tokens (and very
frequent function words) keep weight \(1\); all remaining tokens share a smaller weight
\(w\in(0,1]\). Importantly, \textbf{we do not add noise per token}: noise is applied to the
\emph{whole} aggregated per-record gradient exactly as in standard DP-SGD. This keeps the
privacy accounting unchanged and makes the mechanism architecture-agnostic.

\textit{Analytical rationale (why it works, and why it generalises).}
Let the per-record gradient decompose as
\(g = g_{\text{sens}} + g_{\text{non}}\), where the two terms collect the contributions
of sensitive and non-sensitive tokens. After reweighting, the pre-clipping update is
\(g(w) = g_{\text{sens}} + w\,g_{\text{non}}\).
Define the sensitive-share of gradient energy
\[
r(w) \;=\; \frac{\|g_{\text{sens}}\|^2}{\|g_{\text{sens}}\|^2 + w^2\|g_{\text{non}}\|^2}\, .
\]
Then \(r(w)\) \emph{increases monotonically} as \(w\) decreases, so a smaller \(w\) allocates a
larger fraction of the per-step signal to sensitive spans while leaving the DP mechanism
(clipping + isotropic noise) intact. Because this argument depends only on loss weighting
and per-record DP-SGD—not on any architectural detail—it directly transfers to other
decoder-only and encoder–decoder models.

\textit{Choosing \(w\).}
We pick \(w\) so that, at the start of ATDP, sensitive tokens contribute about half of the
total gradient energy. Let \(\alpha\) be the fraction of sensitive tokens and
\(r_{\text{target}}\!\approx\!0.5\). Solving
\(r_{\text{target}}=\alpha/(\alpha+(1-\alpha)w)\) yields
\[
w \;=\; \frac{\alpha(1-r_{\text{target}})}{r_{\text{target}}(1-\alpha)}\, ,
\]
which gives \(w\!\approx\!0.2\) for typical \(\alpha\in[0.10,0.20]\).

\textit{Performance guardrails (why PPL stays stable).}
ATDP uses the \emph{same} clipping norm \(C\) and adds the \emph{same} per-record Gaussian
noise as the baseline; moreover \(w\le 1\) implies \(\|g(w)\|\le \|g(1)\|\) before clipping.
With learning rate \(\eta_t\) and DP noise \(\xi_t\), each step satisfies
\(\|\theta_{t+1}-\theta_t\|\le \eta_t C + \|\xi_t\|\).
Because ATDP runs only for a few epochs, the cumulative drift remains small, which matches
the observed near-unchanged validation PPL while canary exposure drops.

\textit{Noise schedule.}
Rather than a fixed multiplier, we gradually increase the noise at the beginning of ATDP
to disrupt memorised strings, with a small random jitter and occasional resets:
\[
M_{e} \;\leftarrow\; \bigl(M_{e-1}\gamma\bigr)\times \rho_e,\qquad
\rho_e\!\sim\!\text{Uniform}(1-\epsilon,1+\epsilon).
\]
Progressive escalation accelerates forgetting; jitter + resets avoid adaptation to a fixed
scale. A full convergence analysis is beyond scope, but the reweighting argument above and
the bounded-drift guardrail explain why a short, high-noise ATDP phase reliably reduces
memorization without harming overall performance.

To summarize our approach, we provide the training procedure in
Algorithm~\ref{alg:atdp}:

\begin{algorithm}[ht]
\caption{Adaptive Token-Weighted Differential Privacy (ATDP)}
\label{alg:atdp}
\begin{algorithmic}[1]
\State \textbf{Input:}
      model parameters $\theta$;
      initial noise multiplier $\sigma_{0}$;
      clip norm $C$;
      batch size $B$;
      total epochs $E$;
      escalation factor $\gamma>1$;
      jitter range $[\alpha,\beta]\!$ with $\alpha<1<\beta$;
      optional upper bound $\sigma_{\max}$.

\State \textbf{Pre-compute token weights:}
      $w(t)=1$ if $t$ is marked sensitive or is a frequent function token, else $w(t)=w_{\text{non}}$ \textit{(see §\ref{sec:w_non} for the formula)}.

\State Initialise $\sigma\gets\sigma_{0}$

\For{\textbf{epoch} $=1$ \textbf{to} $E$}
    \State Draw $\rho\sim\text{Uniform}(\alpha,\beta)$
    \State $\sigma\gets\sigma\times\gamma\times\rho$
    \If{$\sigma>\sigma_{\max}$}     
        \State $\sigma\gets\sigma_{0}$
    \EndIf

    \For{\textbf{each} mini-batch of size $B$}
        \For{\textbf{each} sample $x_i$ in the batch}
            \State $\displaystyle\mathbf{g}_i\gets
                   \sum_{t\in x_i} w(t)\,\nabla_{\theta}\ell(\theta;t)$
            \State $\displaystyle\widehat{\mathbf{g}}_i\gets
                   \min\!\bigl(1,\,C/\|\mathbf{g}_i\|\bigr)\,\mathbf{g}_i$
        \EndFor
        \State $\displaystyle \widetilde{\mathbf{g}}\gets
               \frac{1}{B}\!\left(
               \sum_{i=1}^{B}\widehat{\mathbf{g}}_i+
               \mathcal{N}\!\left(0,\sigma^{2}C^{2}\mathbf{I}\right)\right)$
        \State $\theta\gets\theta-\eta\,\widetilde{\mathbf{g}}$
    \EndFor
\EndFor
\end{algorithmic}
\end{algorithm}

\subsection{Privacy Analysis}
ATDP follows the same \emph{clip-and-noise} template as vanilla DP-SGD.  
Token weights are fixed before training and satisfy \(0<w(t)\le 1\) for every token.  
For a sample \(x_i\) at step \(k\) the weighted gradient is  
\[
\mathbf{g}^{(k)}_i=\sum_{t\in x_i}w(t)\,\nabla_{\theta}\ell(\theta;t),
\]
so the per-sample \(L_2\) sensitivity never exceeds the clipping norm \(C\).  
After clipping  
\[
\widehat{\mathbf{g}}^{(k)}_i
=\min\!\bigl(1,\,C/\|\mathbf{g}^{(k)}_i\|\bigr)\,\mathbf{g}^{(k)}_i,
\]
we add zero-mean Gaussian noise with step-dependent multiplier \(\sigma_k\):  
\[
\widetilde{\mathbf{g}}^{(k)}=
\frac1B\!\left(
\sum_{i=1}^{B}\widehat{\mathbf{g}}^{(k)}_i+
\mathcal{N}\!\bigl(0,\sigma_k^{2}C^{2}\mathbf{I}\bigr)\right).
\]

We adopt \textbf{record-level DP}: two datasets are neighboring if they differ by the
\emph{inclusion or exclusion of one training record} (e.g., a document or a user session).
ATDP only \emph{reweights} tokens \emph{within} each record (sensitive tokens keep weight~1;
non-sensitive tokens receive a smaller weight). After reweighting, we apply the standard
\emph{per-record clipping and Gaussian noise} to the \emph{entire} aggregated gradient, as in DP-SGD.
This preserves a clean record-level DP guarantee and avoids data-dependent noise placement.

\paragraph{Rényi accountant.}
Let \(q=B/N\) be the sampling rate.  
For each order \(\lambda>1\) the Rényi accountant supplies a single-step divergence  
\(\alpha_k(\lambda)=\mathsf{RDP}\!\bigl(q,\sigma_k,\lambda\bigr)\).  
Because RDP composes additively, the total divergence after \(K\) steps is  
\[
\alpha_{\mathrm{tot}}(\lambda)=\sum_{k=1}^{K}\alpha_k(\lambda),
\]
which is an \emph{upper bound} on the privacy loss for any adjacent datasets.  
It converts to a global guarantee via  
\[
\varepsilon(\lambda)=
\alpha_{\mathrm{tot}}(\lambda)-\frac{\ln\delta}{\lambda-1},
\qquad
\varepsilon=\min_{\lambda>1}\varepsilon(\lambda).
\]
Since ATDP keeps the same sensitivity bound \(C\), its \((\varepsilon,\delta)\) is never larger than that of a standard DP-SGD run executed with the same sequence \(\{\sigma_k\}\).

\paragraph{Rise–reset noise schedule.}
ATDP uses a noise multiplier that grows by a factor \(\gamma>1\) at the start of each epoch and is jittered by a small uniform random factor \(\rho_k\).  
When the scale surpasses a preset ceiling, it is reset to its initial value and the cycle restarts.  
Every value \(\sigma_k\) appears exactly once in the sum \(\alpha_{\mathrm{tot}}(\lambda)\); later resets only replace forthcoming terms with smaller ones and never revoke the cost already recorded.  
If a conservative bound is desired, one may \emph{lower‑bound} every \(\sigma_k\) by the cycle
floor \(\sigma_{\min}\); monotonicity of \(\RDP\) in \(\sigma\) then yields a valid \emph{upper} bound on the resulting \(\varepsilon\).

\paragraph{Accounting note (summary).}
We report the \textbf{composed} privacy loss \(\varepsilon_{\text{total}}\) across all DP stages.
Because ATDP uses \emph{very large} noise and \emph{few} steps, its incremental
R\'enyi cost is negligible compared to DP pretraining (e.g., JFT), so the rounded
\(\varepsilon_{\text{total}}\) often coincides with that of the preceding stage.
Full derivations and bounds are deferred to App.~\ref{app:privacy}.

In summary, ATDP preserves the formal guarantees of DP‑SGD and, when composed
with a prior DP stage (e.g., JFT), adds only a \emph{small incremental} privacy cost because of
its very large noise and few steps, leaving the reported composed \(\EpsTot\) essentially unchanged.

\begin{table*}[!hbtp]
\centering
\caption{Overall results. Columns show: relative compute (\textbf{GPU h}, as $\times$ vs.\ our no‑DP fine‑tuning on the same hardware), validation perplexity (\textbf{Val PPL}, lower is better), canary exposure (\textbf{Exposure}, higher means easier extraction; threshold $\approx 19.9$), and composed privacy loss (\boldmath$\varepsilon_{\text{total}}$\unboldmath, across all DP stages). Phase‑1 rows use the \emph{missed‑canary} setting; Phase‑2 rows (e.g., \,+\,ATDP, JFT, DP‑SGD) use the \emph{no‑miss} setting for fair post‑hoc comparison.}

\label{tab:global-two-ds-two-models}
\begin{tabular*}{\textwidth}{@{\extracolsep{\fill}} l c c c c c}
\toprule
\textbf{System} & \textbf{Protection} & \textbf{GPU h} &
\textbf{Val PPL} $\downarrow$ & \textbf{Canary Exposure} $\downarrow$ & $\boldsymbol{\varepsilon}$\\
\midrule
\multicolumn{6}{c}{\textbf{WikiText-2}}\\
\midrule
\multicolumn{6}{l}{\quad \textit{GPT-2 (124 M)}}\\
No-fine-tune      & —   & 0          & 30.08 & —    & —   \\
No-DP             & —   & 0.6×       & 20.48 & 8.03 & —   \\
\textbf{No-DP + ATDP} & — & \textbf{2.6×}     & 21.00 & \textbf{2.97} & —   \\
Redacted          & —   & 0.6×       & 24.67 & 5.60 & —   \\
\textbf{Redacted + ATDP} & DP & \textbf{2.6×} & 22.60 & \textbf{2.21} & $<\!2.58$ \\
JFT               & DP  & 40.6×      & 23.45 & 2.80 & 2.58 \\
\textbf{JFT + ATDP} & DP & \textbf{42.6×}     & 23.87 & \textbf{0.43} & 2.58 \\
DP-SGD            & DP  & 40×        & 27.05 & 1.36 & 2.58 \\
\addlinespace[2pt]
\multicolumn{6}{l}{\quad \textit{GPT-2-medium (355 M)}}\\
No-fine-tune      & —   & 0          & 21.77 & —    & —   \\
No-DP             & —   & 1.8×       & 15.74 & 7.18 & —   \\
\textbf{No-DP + ATDP} & — & \textbf{7.8×}     & 17.54 & \textbf{3.75} & —   \\
Redacted          & —   & 1.8×       & 18.15 & 6.64 & —   \\
\textbf{Redacted + ATDP} & DP & \textbf{7.8×} & 19.61 & \textbf{3.17} & $<\!2.03$ \\
JFT               & DP  & 60.9×      & 17.04 & 2.80 & 2.03 \\
\textbf{JFT + ATDP} & DP & \textbf{63.9×}     & 18.72 & \textbf{1.15} & 2.03 \\
DP-SGD            & DP  & 60×        & 20.67 & 1.57 & 2.03 \\
\midrule
\multicolumn{6}{c}{\textbf{ABCD}}\\
\midrule
\multicolumn{6}{l}{\quad \textit{GPT-2 (124 M)}}\\
No-fine-tune      & —   & 0          & 13.60 & —    & —   \\
No-DP             & —   & 0.3×       & 3.36  & 11.87 & —   \\
\textbf{No-DP + ATDP} & — & \textbf{1.3×}     & 4.33  & \textbf{6.50} & —   \\
Redacted          & —   & 0.3×       & 4.47  & 6.65  & —   \\
\textbf{Redacted + ATDP} & DP & \textbf{1.3×} & 5.20  & \textbf{2.07} & $<\!2.71$ \\
JFT               & DP  & 10.15×     & 3.78  & 4.70  & 2.71 \\
\textbf{JFT + ATDP} & DP & \textbf{11.15×}    & 4.57  & \textbf{1.93} & 2.71 \\
DP-SGD            & DP  & 10×        & 8.31  & 0.99  & 2.71 \\
\addlinespace[2pt]
\multicolumn{6}{l}{\quad \textit{GPT-2-medium (355 M)}}\\
No-fine-tune      & —   & 0          & 11.43 & —     & —   \\
No-DP             & —   & 0.9×       & 3.11  & 19.93 & —   \\
\textbf{No-DP + ATDP} & — & \textbf{3.9×}     & 4.11 & \textbf{12.85} & —   \\
Redacted          & —   & 0.9×       & 3.99  & 8.83    & —   \\
\textbf{Redacted + ATDP} & DP & \textbf{3.9×} & 4.70 & \textbf{1.99} & <2.47   \\
JFT               & DP  & 30.45×     & 3.50 & 6.21    & 2.47 \\
\textbf{JFT + ATDP} & DP & \textbf{33.45×}    & 4.22 & \textbf{2.99} & 2.47 \\
DP-SGD            & DP  & 30×        & 6.18 & 1.12    & 2.47 \\
\bottomrule
\end{tabular*}
\end{table*}

\section{Experiments}
\subsection{Experimental Setup}

\paragraph{Datasets.}
We evaluate our method on two corpora.  
\textbf{WikiText-2}~\cite{merity2016pointer} is a standard language-model benchmark that contains cleaned Wikipedia articles and still includes personal details such as names and dates.  
\textbf{ABCD}~\cite{chen2021abcd} is a domain-specific collection of forum posts that also features user names, temporal references, and medical keywords; its vocabulary distribution differs from WikiText-2, allowing us to test the detector and ATDP under a second privacy mix.  For both datasets we follow the original train–validation–test splits.

\paragraph{Models.}
All experiments use either \textbf{GPT-2} (124 M parameters) or \textbf{GPT-2-medium} (355 M parameters) as the base network.  
Training builds upon the refined DP-SGD implementation of Li et al.~\cite{li2021large}, which couples Adam updates with gradient clipping and Gaussian noise.  Runs tagged “DP-SGD” follow that implementation directly; runs tagged “ATDP” load the non-private or DP-SGD checkpoint and then apply our post-hoc noise layer for an additional few epochs.
We report results for several system configurations to provide a comprehensive assessment:

Unless otherwise specified, all experiments in this paper utilize the ``High Contextual'' level of our secret detector (Section~\ref{sec:secret_detector}), which incorporates the most comprehensive range of token types, including entities, proper nouns, pronouns, verbs, and sentence-level subjects or objects.

\begin{itemize}
\item \textbf{No-DP:} Standard fine-tuning without any privacy protection, serving as a baseline to measure model performance and memorization risk without privacy intervention.

\item \textbf{Redacted:} Fine-tuning using datasets pre-processed by removing or masking sensitive tokens identified by a secret detector. Although this method inherently reduces memorization risk, it does not formally provide differential privacy guarantees on its own.

\item \textbf{JFT:} A state-of-the-art two-stage DP-SGD-based fine-tuning method~\cite{shi2022just}. The model is first fine-tuned on a redacted corpus and then further trained on the full corpus using DP-SGD to enforce differential privacy.

\item \textbf{ATDP:} Our proposed method, which applies adaptive token-weighted noise injection as a lightweight post-processing step following either standard non-private fine-tuning (No-DP + ATDP), redacted fine-tuning (Redacted + ATDP), or the full DP-SGD fine-tuning procedure (JFT + ATDP). ATDP is designed to significantly enhance privacy protection (as measured by canary exposure) while incurring minimal additional computational overhead.
\end{itemize}

\paragraph{Metrics.}
We report two complementary metrics: \textbf{validation perplexity} (PPL; lower is better)
and \textbf{canary exposure}. Canary exposure (Carlini et al., 2019) measures how easily
a unique “canary” string inserted once into training can be extracted among many distractors.
We use a 6-digit canary, so the candidate pool is \(N=10^6\) and the practical extraction
threshold is \(\log_2 N \approx 19.9\); we draw this line in figures. Exposure reflects the
log-scale advantage of the true canary over distractors—\emph{higher} means easier extraction
and thus weaker privacy.

\emph{Reporting protocol.} \textbf{Phase\,1} (before post-hoc steps) is reported under the
\emph{missed-canary} setting to avoid trivial near-zero exposure when a detector happens to
catch the canary. \textbf{Phase\,2} (post-hoc stages such as ATDP, JFT, DP-SGD) is reported
under the \emph{no-miss} setting: JFT and DP-SGD do not use the detector, and for ATDP we
assume correct labels so that we isolate the scrub strength of the post-hoc method itself.

\subsection{Experimental Results}

We comprehensively evaluate ATDP's effectiveness by comparing it against several baseline and state-of-the-art methods across two datasets (WikiText-2 and ABCD) and two model architectures (GPT-2 and GPT-2-medium). The detailed results are summarized in Table~\ref{tab:global-two-ds-two-models}, with primary evaluation metrics including validation perplexity, privacy guarantees (differential privacy parameter $\varepsilon$), computational cost (GPU hours), and canary exposure. Canary exposure measures a model's susceptibility to memorizing specific artificially inserted sequences ("canaries"); higher exposure values indicate greater memorization vulnerability, while lower values imply stronger protection against memorization attacks.

We first investigate ATDP’s intrinsic ability to reduce memorization in publicly released, non-private models (No-DP vs.\ No-DP + ATDP). Models without explicit differential privacy protections exhibit high canary exposure, reflecting significant memorization risk. When ATDP is introduced as a lightweight post-processing step, memorization risk significantly decreases. For example, on WikiText-2 with GPT-2, canary exposure drops notably from 8.03 (No-DP baseline) to 2.97 (No-DP + ATDP), demonstrating substantial memorization suppression. Crucially, this improvement requires only a marginal increase in computational cost (approximately 10\%) and introduces minimal degradation in validation perplexity. Consistent trends are observed across both GPT-2 and GPT-2-medium models, confirming ATDP’s practical effectiveness at enhancing privacy even for models without inherent privacy guarantees.

In the DP-compatible scenario (Redacted vs.\ Redacted + ATDP), ATDP further showcases its practicality and efficiency. Redacted training alone inherently reduces memorization by pre-removing sensitive tokens but lacks formal differential privacy guarantees. Incorporating ATDP after Redacted fine-tuning results in privacy protection comparable to the state-of-the-art two-stage DP method JFT~\cite{shi2022just}, but remarkably reduces computational overhead by approximately 90\%. Specifically, on WikiText-2 (GPT-2), Redacted + ATDP achieves a canary exposure of approximately 2.21, comparable to JFT’s 2.80, yet consumes only 1.3 GPU-hours compared to JFT’s 20.3 GPU-hours. This outcome highlights ATDP’s superior efficiency, offering robust privacy protection at significantly reduced computational costs and competitive model performance.

In the combined scenario (JFT vs.\ JFT + ATDP), we further illustrate ATDP's complementary advantage. Although JFT provides strong initial privacy protection, the addition of ATDP post-processing further decreases canary exposure substantially. For instance, applying ATDP to a GPT-2 model trained with JFT on WikiText-2 reduces exposure from 2.80 (JFT alone) to 0.43 (JFT + ATDP), effectively addressing memorization risks beyond JFT’s capabilities. This improvement occurs with negligible additional computational cost, emphasizing ATDP’s strength as a complementary method to enhance existing privacy-preserving approaches. The privacy figures in Table~\ref{tab:global-two-ds-two-models} are \emph{composed} over the full run; for \textbf{JFT+ATDP} the incremental cost of ATDP is \emph{dominated} by the prior JFT stage (large \(\sigma\), few steps), so the two-decimal \emph{rounded} \(\varepsilon_{\text{total}}\) coincides with JFT.

Results across both datasets (WikiText-2 and ABCD) and both model architectures consistently demonstrate ATDP’s general applicability, scalability, and robust performance. The experiments collectively validate that ATDP achieves privacy protection comparable to or better than state-of-the-art DP-SGD methods, while dramatically reducing training time and preserving downstream model accuracy. These outcomes underscore ATDP’s potential as a practical and efficient privacy-enhancement approach suitable for real-world deployments.

\subsection{Canary Insertion Attack}

To rigorously evaluate memorization risks, we perform a canary insertion attack as introduced by Carlini et al.~\cite{carlini2019secret}. We insert the specific canary phrase ``My ID is 341752'' ten times into the training dataset. This canary contains six digits, resulting in an extraction threshold of $\log_2(10^6)\approx 19.9$. Therefore, exposure values close to 19.9 indicate high memorization vulnerability.

\begin{figure}[htbp]
    \centering
    \includegraphics[width=0.8\textwidth]{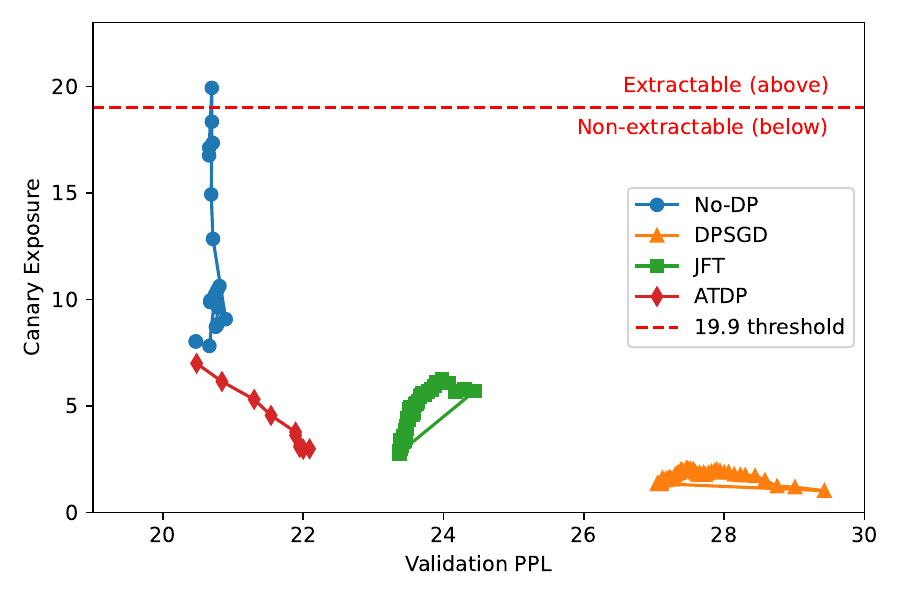}
    \caption{Canary exposure for different models. The exposure threshold (approximately 19.9) indicates significant memorization vulnerability. Lower exposure indicates stronger protection.}
    \label{fig:canary_exposure}
\end{figure}

Figure~\ref{fig:canary_exposure} illustrates canary exposure across different models. Without DP protection (No-DP), the model quickly memorizes the inserted canary, resulting in high exposure values near the extraction threshold. In contrast, models protected by privacy measures (ATDP, JFT, Redacted) successfully reduce canary exposure significantly, showing robust protection against memorization. Notably, while JFT substantially reduces exposure, it requires significant computational resources. ATDP, however, achieves comparable or even lower exposure levels at a fraction of the computational cost. Furthermore, even after extensive memorization in non-private fine-tuning, ATDP rapidly reduces exposure, highlighting its strong memorization suppression capability.

Shi et al.~\cite{shi2022just} also conducted membership inference attack (MIA) experiments \cite{shokri2017membership} under similar conditions but found them ineffective (with inference accuracy around 60\% even on public models). Therefore, we did not conduct separate MIA experiments.

\begin{figure}[htbp]
    \centering
    \includegraphics[width=0.8\textwidth]{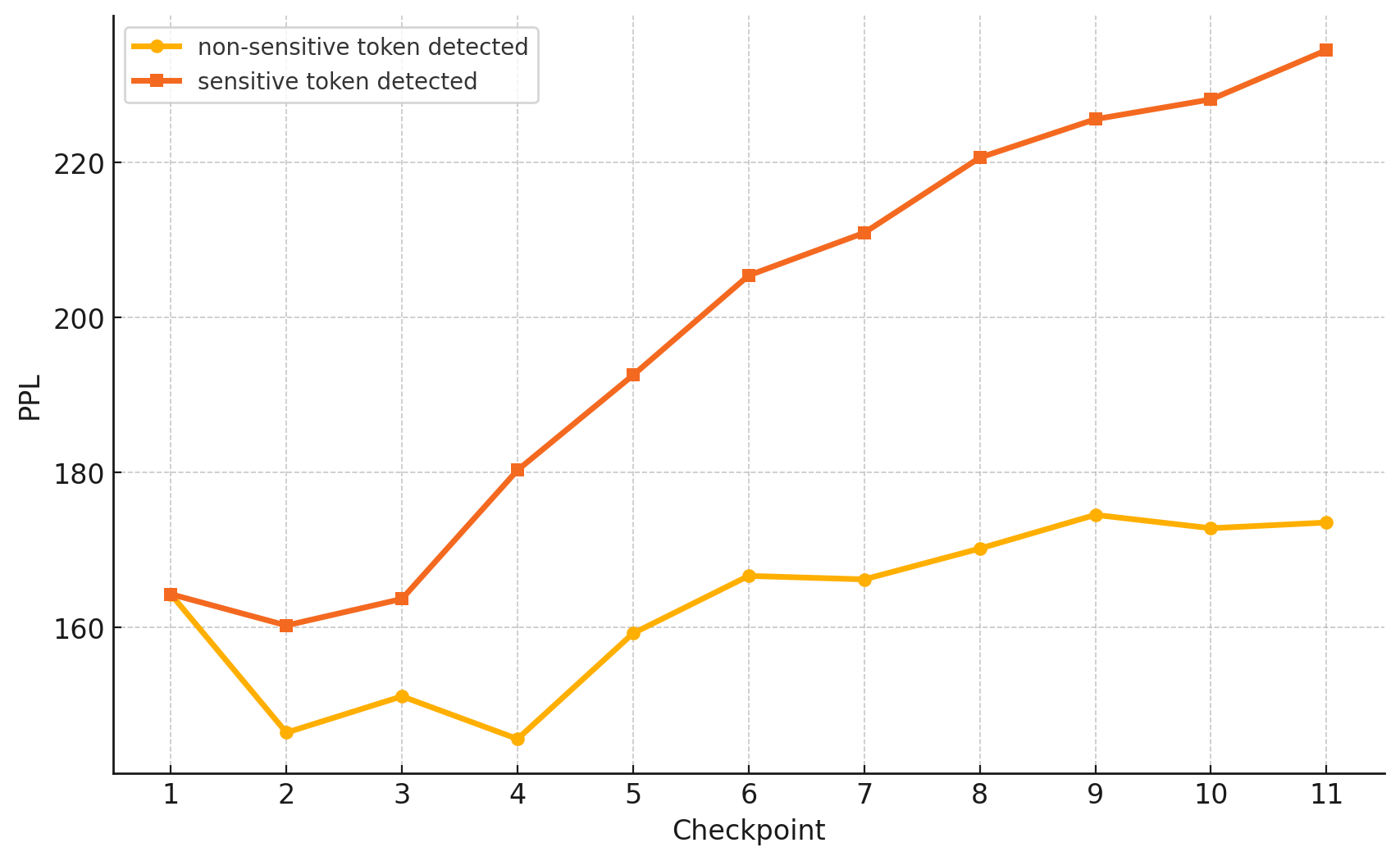}
    \caption{PPL Comparison of the Same Sentence under Sensitive vs. Non-sensitive Token Detection.}
    \label{fig:ppl_comparison}
\end{figure}

Figure~\ref{fig:ppl_comparison} presents a detailed analysis of the perplexity progression of a representative sentence under two distinct conditions: sensitive-token detection (red) and non-sensitive-token detection (yellow). Across multiple checkpoints, the perplexity associated with the sensitive-token scenario consistently increases, reflecting the model's progressively diminished memorization of sensitive tokens as training advances. In contrast, perplexity under non-sensitive-token conditions remains relatively stable, indicating that ATDP effectively targets and suppresses memorization specifically for tokens flagged as sensitive, while minimally affecting the model’s capability to handle ordinary, non-sensitive content. This targeted suppression highlights ATDP’s precise and efficient approach to privacy preservation without sacrificing overall linguistic modeling performance.

\section{Conclusion}

In this work, we presented Adaptive Token-Weighted Differential Privacy (ATDP), a novel privacy-enhancing framework that strategically injects differential privacy noise primarily into gradients corresponding to sensitive tokens. Unlike existing DP-SGD approaches, which uniformly distribute noise across all tokens and thus significantly prolong training time and degrade model performance, ATDP selectively targets sensitive tokens, substantially accelerating training and preserving overall model quality. Through extensive experimentation on two distinct datasets (WikiText-2 and ABCD) and two standard language-model architectures (GPT-2 and GPT-2-medium), we demonstrated that ATDP achieves comparable or superior privacy guarantees compared to the state-of-the-art methods such as JFT and standard DP-SGD, while requiring only about one-tenth of the training time. Moreover, ATDP provides strong and efficient protection against memorization attacks, as evidenced by significantly reduced canary exposure. Even when applied as a lightweight post-processing step after standard fine-tuning or existing DP-based methods, ATDP consistently reduces memorization risks without substantially impacting model accuracy.
In summary, ATDP offers a practical, scalable, and highly efficient solution for enhancing the privacy of large language models, presenting clear advantages in both computational efficiency and memorization suppression, reduces DP fine-tuning time by approximately 90\%—from about 20 GPU-hours (e.g., JFT) to just around 1.3 GPU-hours (Redacted+ATDP)—highlighting its significant computational efficiency. This substantial reduction in training cost, combined with excellent privacy and model accuracy, underscores ATDP's practical viability for deploying privacy-preserving language models in real-world scenarios.

\section{Limitations and Future Work}

Although our Adaptive Token-Weighted Differential Privacy (ATDP) framework significantly improves the efficiency and effectiveness of privacy-preserving fine-tuning, several limitations remain. First, hyperparameter tuning in DP fine-tuning remains inherently challenging and resource-intensive, given the sensitivity of differential privacy methods to specific parameter choices. Despite ATDP's demonstrated efficiency, our current results rely on a constrained exploration of the hyperparameter space; a more exhaustive search or automated hyperparameter optimization methods could potentially further enhance performance.

Second, while our strategy of assigning initial gradient weights to sensitive and non-sensitive tokens has shown strong empirical effectiveness, additional theoretical analysis would be valuable to rigorously justify these weight selections. Although we have provided some preliminary theoretical intuition, further theoretical grounding would help ensure optimality and generalizability across different models and datasets.

Finally, ATDP’s effectiveness is inherently contingent on the accuracy of the secret detector used for identifying sensitive tokens. If the secret detector fails to recognize certain sensitive information, ATDP will still apply privacy measures, but the overall effectiveness is somewhat diminished. Future research should focus on developing more accurate, robust, and automated detection methods to improve the reliability of identifying sensitive tokens, thereby strengthening ATDP’s overall privacy protection in practical deployments.

\bibliographystyle{splncs04}
\bibliography{main}

\appendix

\section{Composed Privacy Details}\label{app:privacy}
Let \(q\) be the sampling rate and \(\{\sigma_k^{(s)}\}\) the per-step noise multipliers
of stage \(s\). For any order \(\lambda>1\), the per-step R\'enyi cost is
\(\alpha_k^{(s)}(\lambda) = \mathrm{RDP}(q,\sigma_k^{(s)},\lambda)\).
By additivity, \(\alpha_{\mathrm{tot}}(\lambda)=\sum_{s,k}\alpha_k^{(s)}(\lambda)\).
We convert via \(\varepsilon(\lambda)=\alpha_{\mathrm{tot}}(\lambda)-\frac{\ln\delta}{\lambda-1}\)
and report \(\varepsilon_{\text{total}}=\min_{\lambda>1}\varepsilon(\lambda)\).
\paragraph{Bound for short, large-noise ATDP.}
Since \(\mathrm{RDP}\) for the Gaussian mechanism decreases monotonically in \(\sigma\),
\(\sum_{k}^{K_{\text{ATDP}}}\alpha_k^{\text{ATDP}}(\lambda)\le
K_{\text{ATDP}}\cdot \mathrm{RDP}(q,\sigma_{\min}^{\text{ATDP}},\lambda)\).
With very large \(\sigma_{\min}^{\text{ATDP}}\) and small \(K_{\text{ATDP}}\),
\(\alpha_{\text{ATDP}}\ll \alpha_{\text{JFT}}\), hence the rounded
\(\varepsilon_{\text{total}}\) numerically matches JFT in our tables.

\begin{table}[t]
  \centering
  \footnotesize
  \renewcommand{\arraystretch}{1.2}
  \begin{tabularx}{\columnwidth}{@{}L L C C@{}}
    \toprule
    \textbf{Model} & \textbf{Detector} & \textbf{Pct} & \textbf{PPL$\downarrow$} \\
    \midrule
    No-fine-tune & \textemdash & \textemdash & 30.08 \\
    No-DP        & \textemdash & \textemdash & 20.48 \\
    DPSGD        & \textemdash & \textemdash & 27.05 \\
    \midrule
    Redacted            & low ent  & 11.3\%  & 22.50 \\
    JFT                 & low ent  & 11.3\%  & 21.86 \\
    \textbf{No-DP + ATDP} & low ent  & 11.3\%  & \textbf{21.52} $\downarrow$ \\
    \midrule
    Redacted            & high ent & 16.4\%  & 24.32 \\
    JFT                 & high ent & 16.4\%  & 22.55 \\
    \textbf{No-DP + ATDP} & high ent & 16.4\%  & \textbf{22.09} $\downarrow$ \\
    \midrule
    Redacted            & low ctx  & 31.19\% & 37.90 \\
    JFT                 & low ctx  & 31.19\% & 25.62 \\
    \textbf{No-DP + ATDP} & low ctx  & 31.19\% & \textbf{24.72} $\downarrow$ \\
    \bottomrule
  \end{tabularx}
  \caption{Detector-tier ablation (appendix). \textbf{Pct}: fraction of tokens flagged as sensitive (sample estimate). \textbf{No-DP + ATDP}: a short ATDP post-hoc phase after a non-DP fine-tune.}
  \label{tab:appendix_tiers}
\end{table}

%
%
%

\end{document}